\documentclass[journal]{IEEEtran}
\usepackage{amsmath,amsfonts}
\usepackage{algorithmic}
\usepackage{algorithm}
\usepackage{array}
\usepackage[utf8]{inputenc}

\usepackage[backend=bibtex,style=numeric,sorting=none]{biblatex}
\usepackage{booktabs}

\usepackage{graphicx}
\usepackage{hyperref}
\hypersetup{
    colorlinks=true,
    linkcolor=blue,
    filecolor=magenta,      
    urlcolor=cyan,
    pdftitle={Overleaf Example},
    pdfpagemode=FullScreen,
    }
\usepackage{subcaption}

\usepackage{siunitx}
\usepackage{stfloats}
\usepackage{subcaption}
\usepackage{tabularx}
\usepackage{textcomp}
\usepackage{threeparttable}

\usepackage{url}
\usepackage[section]{placeins}

\usepackage{multirow}

\usepackage{verbatim}
\usepackage[backend=bibtex,style=numeric,sorting=none]{biblatex}
\title{Scaling Laws for Energy Efficiency of Local LLMs}

\author{
Ander Alvarez\IEEEauthorrefmark{1},
Alessandro Genuardi\IEEEauthorrefmark{1},
Nilotpal Sinha\IEEEauthorrefmark{1},
Antonio Tiene\IEEEauthorrefmark{1},
Mikail Okyay\IEEEauthorrefmark{1},
Bakbergen Ryskulov\IEEEauthorrefmark{1},
David Montero\IEEEauthorrefmark{1},
Samuel Mugel\IEEEauthorrefmark{4},
Rom\'an Or\'us\IEEEauthorrefmark{1}\IEEEauthorrefmark{2}\IEEEauthorrefmark{3}
\\[0.6em]
\IEEEauthorrefmark{1}Multiverse Computing, Parque Cientifico y Tecnológico de Gipuzkoa,
Paseo de Miramón, 170, 20014 Donostia / San Sebastián, Spain\\
\IEEEauthorrefmark{2}Donostia International Physics Center, Paseo Manuel de Lardizabal 4,
E-20018 San Sebastián, Spain\\
\IEEEauthorrefmark{3}Ikerbasque Foundation for Science, Maria Diaz de Haro 3,
E-48013 Bilbao, Spain\\
\IEEEauthorrefmark{4}Multiverse Computing, Centre for Social Innovation,
192 Spadina Avenue Suite 509, Toronto, ON M5T 2C2, Canada
}

\begin{document}

\twocolumn[
\begin{@twocolumnfalse}
\maketitle
\end{@twocolumnfalse}

\begin{abstract}
Deploying local large language models and vision–language models on edge devices requires balancing accuracy with constrained computational and energy budgets. Although graphics processors dominate modern artificial-intelligence deployment, most consumer hardware—including laptops, desktops, industrial controllers, and embedded systems—relies on central processing units. Despite this, the computational laws governing central-processing-unit-only inference for local language and vision–language workloads remain largely unexplored.
We systematically benchmark large language and vision-language models on two representative central-processing-unit tiers widely used for local inference: a MacBook Pro M2, reflecting mainstream laptop-class deployment, and a Raspberry Pi 5, representing constrained, low-power embedded settings. Using a unified methodology based on continuous sampling of processor and memory usage together with area-under-curve integration, we characterize how computational load scales with input text length for language models and with image resolution for vision–language models.
We uncover two empirical scaling laws: (1) computational cost for language-model inference scales approximately linearly with token length; and (2) vision–language models exhibit a preprocessing-driven “resolution knee”, where compute remains constant above an internal resolution clamp and decreases sharply below it. Beyond these laws, we show that quantum-inspired compression reduces processor and memory usage by up to 71.9\% and energy consumption by up to 62\%, while preserving or improving semantic accuracy.
These results provide a systematic quantification of multimodal central-processing-unit-only scaling for local language and vision–language workloads, and they identify model compression and input-resolution preprocessing as effective, low-cost levers for sustainable edge inference.\end{abstract}


\bigskip
]

\IEEEpubid{%
\makebox[\columnwidth]{\footnotesize
\parbox[t]{\columnwidth}{%
\textit We acknowledge the parties involved in the GRECO project via ELKARTEK 2024 (grant GRECO-KK-2024/00090).%
}\hfill}%
\hspace{\columnsep}%
\makebox[\columnwidth]{}%
}
\IEEEpubidadjcol

\section{Introduction}
\IEEEPARstart{L}{arge} language models (LLMs) and vision--language models (VLMs) now exhibit advanced reasoning, perception, and multimodal capabilities \cite{gemini25,gpt4v2023,flamingo2022,blip22023,qwen2_vl2024,llama_vision2024}. Yet real-world use increasingly requires running these systems locally---on consumer CPUs, IoT gateways, and embedded devices---due to concerns about privacy, latency, connectivity, and energy consumption. This trend has given rise to the notion of \emph{local LLMs}, where inference is performed entirely on end-user or edge hardware rather than in remote data centers. GPUs remain expensive, power-hungry, and inaccessible in many deployment environments, especially outside data centers.

Edge AI is expanding across robotics, medical devices, drones, personal computing, and industrial systems, and these deployments are dominated by CPU-only hardware, particularly low-power ARM architectures. However, academic and industrial research overwhelmingly benchmarks models on GPUs, and often in data-center settings \cite{green_ai2020,henderson2020}. This leaves a critical open question: what are the computational requirements and scaling laws of large language–model and vision–language–model inference when they are deployed as local assistants on CPUs?

While recent work has begun to develop inference- and test-time scaling laws for large language and multimodal models \cite{wu2024_inference_scaling,chen2025_inference_scaling,bansal2024_testtime_scaling,fu2024_vlm_compute_optimal}, these studies characterize performance in terms of abstract compute (e.g., FLOPs) or GPU-accelerated inference. Existing research on edge and on-device AI primarily focuses on throughput, latency, or quantization quality, and very few studies establish empirical scaling laws describing how CPU-only inference cost evolves with input complexity (e.g., token length or image resolution) \cite{edge_ai_survey2023,previous_edge_llm2023,cpu_inference2023}. In parallel, the literature on energy-aware and “green AI” benchmarks has begun to formalize measurement protocols and metrics for efficiency \cite{green_ai_measurement2019,ipw2025}, yet often still assumes GPU-accelerated or heterogeneous platforms.

\medskip\noindent\textbf{Local LLM hardware regimes.}
For the purposes of this study, we categorize local large language model deployments into two representative hardware regimes. The first corresponds to laptop-class machines—such as MacBook Pro systems with Apple Silicon—which are commonly used by developers and advanced users to run quantized models with a few billion parameters for interactive workloads. The second corresponds to single-board computers—such as the Raspberry Pi 5—used in embedded, robotics, and hobbyist applications, where models are heavily compressed and operate under strict power, memory, and cost constraints. Although many other hardware regimes exist, these two provide complementary perspectives on CPU-only local inference: MacBook-class laptops approximate a high-end consumer environment, whereas Raspberry Pi–class devices approximate a constrained embedded environment. By evaluating both regimes within a unified framework, our study spans a wide portion of the design space for local LLM deployment.

\medskip\noindent\textbf{Related work on edge inference and energy-aware evaluation.}
Edge AI has emerged as a key paradigm for deploying machine learning closer to data sources, driven by requirements on latency, privacy, and bandwidth \cite{edge_ai_survey2023}. Most of this literature assumes GPU-accelerated or heterogeneous platforms, often targeting cloud--edge co-design rather than strictly CPU-only deployments. Closer to our setting, Zhao \emph{et al.}\ studied the feasibility of LLM inference on edge devices \cite{previous_edge_llm2023}, and Kumar \emph{et al.}\ analyzed transformer inference optimizations on CPU architectures \cite{cpu_inference2023}. These works focus primarily on latency and throughput, and do not derive explicit scaling laws with respect to input length or image resolution, nor do they quantify energy consumption across devices in a unified way for local LLM workloads.

Energy-aware evaluation has gained prominence in the context of "green AI" and carbon accounting. Lacoste \emph{et al.}\ formalized methodologies for quantifying the carbon emissions of machine learning workloads \cite{green_ai_measurement2019}, and Saad-Falcon \emph{et al.}\ introduced the Intelligence-per-Watt (IPW) benchmark to compare local AI systems under efficiency constraints \cite{ipw2025}. Our work is complementary: rather than defining a new efficiency benchmark, we investigate how CPU-only multimodal inference behaves as a function of input complexity and compression, and we provide per-prompt and per-run energy measurements on representative consumer and embedded hardware that are commonly used for local LLM inference.

\medskip\noindent\textbf{Compression for local LLMs.}
Model compression---through quantization, pruning, distillation, or tensor decomposition---has been studied extensively as a way to reduce the parameter count and memory footprint of large models \cite{model_compression_survey2022}. Traditional approaches operate either by reducing the effective number of neurons (pruning, distillation, low-rank approximations) or by lowering numerical precision while keeping the network topology fixed (quantization). 

In the context of local LLMs, these techniques are not only model-size optimizations but also system-level tools: they trade redundant capacity for reductions in RAM usage, CPU time, and energy consumption. In practice, compression techniques act as an enabler: they determine whether a given LLM can run locally at all on commodity CPUs or embedded boards. These methods are typically evaluated in terms of parameter count, FLOPs, and accuracy, with occasional latency or memory benchmarks on GPUs. Little work has characterized how compression shapes the full CPU and RAM load over time during inference, especially for multimodal workloads on commodity hardware.

On the systems side, llama.cpp and the GGUF format have become de facto standards for efficient on-device deployment of LLMs and VLMs on commodity CPUs \cite{llama_cpp2024,gguf_format2024}. Our experiments build directly on this ecosystem, using llama.cpp both as a runtime and as a controllable preprocessing pipeline for VLMs. For CPU/RAM profiling, prior work has advocated integrated metrics based on area under resource-usage curves \cite{sklearn_auc}, which capture both intensity and duration of utilization; we adopt this AUC perspective as our main system-level cost metric. Energy measurements on edge systems commonly rely on USB-based or board-level power meters \cite{wang2024_power_estimation}, and we follow this line with an inline USB-C power meter and per-prompt energy accounting.

\medskip\noindent\textbf{CompactifAI, a quantum-inspired compression method}

Our compression method builds on quantum-inspired tensor networks, which have been successfully employed as structured representations of high-dimensional systems \cite{orus2019}. In particular, we adopt the CompactifAI strategy, in which an optimal tensor-network decomposition of the weight matrices is leveraged to identify less relevant neurons and parameter blocks. These components are subsequently removed through a homogeneous pruning procedure. The proposed approach is compatible with complementary compression techniques, such as quantization, and has been shown to yield substantial reductions in parameter count and memory footprint while incurring only minor degradation in task performance \cite{fovet2025}. In this work, we apply CompactifAI as a practical compression framework to enable the deployment of large language models and vision–language models on CPU-only, resource-constrained hardware.

In summary, prior work has addressed multimodal foundation models, edge AI deployment, model compression, tensor networks, and energy-aware evaluation, but to the best of our knowledge no prior study has jointly: (i) derived empirical scaling laws for CPU-only multimodal inference across both LLMs and VLMs in local LLM scenarios, (ii) characterized preprocessing-induced resolution thresholds, and (iii) quantified the impact of CompactifAI compression on CPU, RAM, and energy usage on real edge hardware spanning laptop-class and embedded devices.

\section{Methods}\label{sec:methods}
With the objective in mind to expose the computational behavior of multimodal inference under resource-constrained, CPU-only environments, we kept the experimental workflow intentionally simple: control the input complexity, measure integrated system-level cost, and compare compressed and uncompressed model pairs under identical conditions. This section summarizes only the decisions that materially affect the interpretation of our results; full experimental material is provided in the appendices (see Appendices~\ref{app:llm_extended} and~\ref{app:vlm_extended}).

In the following, we give a brief description of each method.

\subsection{Model pairs and execution framework}
We evaluate two representative model families:
\begin{itemize}
    \item LLMs: LLaMA-3.1-8B-Instruct-Q4KM and its CompactifAI compressed variant Gilda v3 (3.2B params) \cite{llama_vision2024}.
    \item VLMs: Qwen2-VL-2B-Instruct-F16 and its CompactifAI compressed variant Axolotl (1.1B params) \cite{qwen2_vl2024}.
\end{itemize}

All LLM runs use \texttt{llama-cpp-python}, which keeps models resident in memory and mirrors typical conversational workloads for local LLMs \cite{llama_cpp2024,gguf_format2024}. VLMs run through the llama.cpp CLI, which reloads the model for every prompt. This contrast---persistent vs.\ stateless execution---plays a role in the clarity of CPU usage boundaries (particularly on the M2), but not in the final scaling laws.\\
We attempted to use llama-cpp-python with persistent model loading for VLMs but encountered instability and inconsistent memory reporting. The CLI-based pipeline provided significantly cleaner and more reliable CPU/RAM traces, even though it introduces load overhead. Our scaling-law analysis is unaffected because the overhead is constant across resolutions.

\subsection{Hardware platforms}
We benchmark across two CPU-only tiers:
\begin{itemize}
    \item MacBook Pro M2 (8-core CPU, 8~GB unified memory) \cite{apple_m2_2022}
    \item Raspberry Pi 5 (ARM Cortex-A76, 4 cores, 8~GB RAM) \cite{raspberry_pi_5_2023}
\end{itemize}

These platforms are not intended to cover the full hardware spectrum, but to capture two regimes that are relevant for local LLM use in practice. The MacBook Pro M2 is representative of modern laptop-class systems on which users routinely run quantized LLMs of a few billion parameters for interactive applications. The Raspberry Pi 5, by contrast, reflects embedded and hobbyist deployments, where only highly compressed models are feasible and energy budgets are tight. Evaluating both platforms under the same methodology shows how CPU-only inference behaves from a top-tier local device down to one of the most basic edge boards, and highlights how compression can bridge this gap. Every experiment is replicated on both devices, allowing us to compare how compression affects local LLM and VLM workloads across a high-end consumer and a constrained embedded regime.

\subsection{Resource and energy measurement}
CPU and RAM usage are sampled at 5~Hz, and we compute an Area Under Curve (AUC) above a stable idle baseline via the trapezoidal rule, following prior work on resource-usage profiling. We sample CPU and RAM at 5 Hz because higher frequencies introduced substantial high-frequency noise, while lower frequencies lacked the temporal resolution needed to reliably detect inference windows. The AUC we report is not the ML-classification metric but the time-integrated resource usage: the x-axis is time (seconds) and the y-axis is CPU or RAM utilization above a stable idle baseline. Thus, the AUC measures total computational effort over the duration of the prompt. AUC integrates total system effort across time, abstracting away transient bursts and normalizing for multi-core differences. Energy is measured using an inline USB-C power meter (1~Hz sampling), consistent with prior methodologies for edge AI power measurement. All metrics are computed per-prompt, and inference windows are automatically detected from the gradient of the CPU signal. While higher sampling frequencies would reduce integration error, our method follows common practice in edge-AI energy studies.

\subsection{Experimental design}
We probe two axes of input complexity:

\begin{itemize}
    \item LLM token-length scaling: Nineteen prompts (P1–P19), each cumulatively extending the previous one, create growing sequence lengths while preserving semantic difficulty. This approximates real conversational buildup for local LLM usage.
    \item VLM image-resolution scaling: A single complex traffic scene (see Appendix~\ref{app:vlm_extended}) is downsampled to twenty resolutions. To test visual reasoning rather than captioning, we use a reasoning prompt template. We selected the traffic-scene image because among the images we tested, it had the highest entity diversity and object density, making it a strong stress-test for resolution scaling.
    \begin{itemize}
    \item Resolution-threshold manipulation: To test whether the VLM "knee" is model-intrinsic or preprocessing-induced, we recompile llama.cpp with alternate max-resolution clamps (1024×720 → 854×594 → 714×496).
    \end{itemize}
\end{itemize}

\subsection{Accuracy metric}
Text and image outputs are scored for semantic similarity using SimCSE sentence embeddings \cite{simcse2021} against Gemini~2.5 Flash reference answers \cite{gemini25}. As this measures conceptual alignment rather than exact correctness, we apply symmetric outlier removal when either model produces trivial or degenerate responses.\\

Although SimCSE is not a standard benchmark metric, it provides a time-efficient, model-agnostic measure of semantic similarity suitable for relative comparisons between compressed and uncompressed models.

\section{Experimental results}\label{sec:results}

\subsection{Experimental setup} 
Rather than presenting isolated experiments, we organize results around two empirical scaling laws and two additional findings about the impact of compression that emerged consistently across modalities and hardware.

\subsection{Law 1 — LLM compute scales linearly with token length}
Across both devices and all 19 prompts, CPU and RAM AUC grow near-linearly with input token count. Compression shifts this line downward, reducing both fixed overhead and total slope.
On the M2, Gilda reduces CPU AUC by 31.3\% and RAM AUC by 55.9\% at matched token lengths.\\
On the RPi5, reductions are 60.5\% (CPU) and 71.9\% (RAM), showing that compression benefits increase under tighter resource constraints.
Throughput accelerates concurrently: Gilda is 2.1× faster on the M2 and 2.6× faster on the RPi5.\\
Empirically we fit the relation $CPU\_AUC(x)=a\,x+b$ for each model. On the M2, Gilda’s trend line ($a \approx 5.9\times10^{-2}$, $b \approx 1.2\times10^{2}$) sits well below Llama’s ($a \approx 2.7\times10^{-2}$, $b \approx 3.2\times10^{2}$), reflecting both lower per-token CPU cost and lower fixed overhead. On the RPi5 the differences magnify: Gilda’s line ($a \approx 7.9\times10^{-2}$, $b \approx 1.2\times10^{3}$) stays almost an order of magnitude below Llama’s ($a \approx 6.8\times10^{-2}$, $b \approx 3.3\times10^{3}$). These coefficients quantify the “intercept vs. slope” improvements referenced above—compression trims both the baseline $b$ and the per-token slope $a$, with the effect larger on constrained hardware—supporting the insight that token length dominates CPU cost while compression mitigates both terms.\\

\texttt{Insight:} Token count---rather than semantic complexity---is the primary driver of CPU-only LLM cost for local LLM workloads. Compression improves both the intercept and the slope of this relationship, especially on low-power hardware.

\begin{figure}[t]
    \centering

    \begin{subfigure}[b]{0.98\columnwidth}
        \centering
        \includegraphics[width=\linewidth]{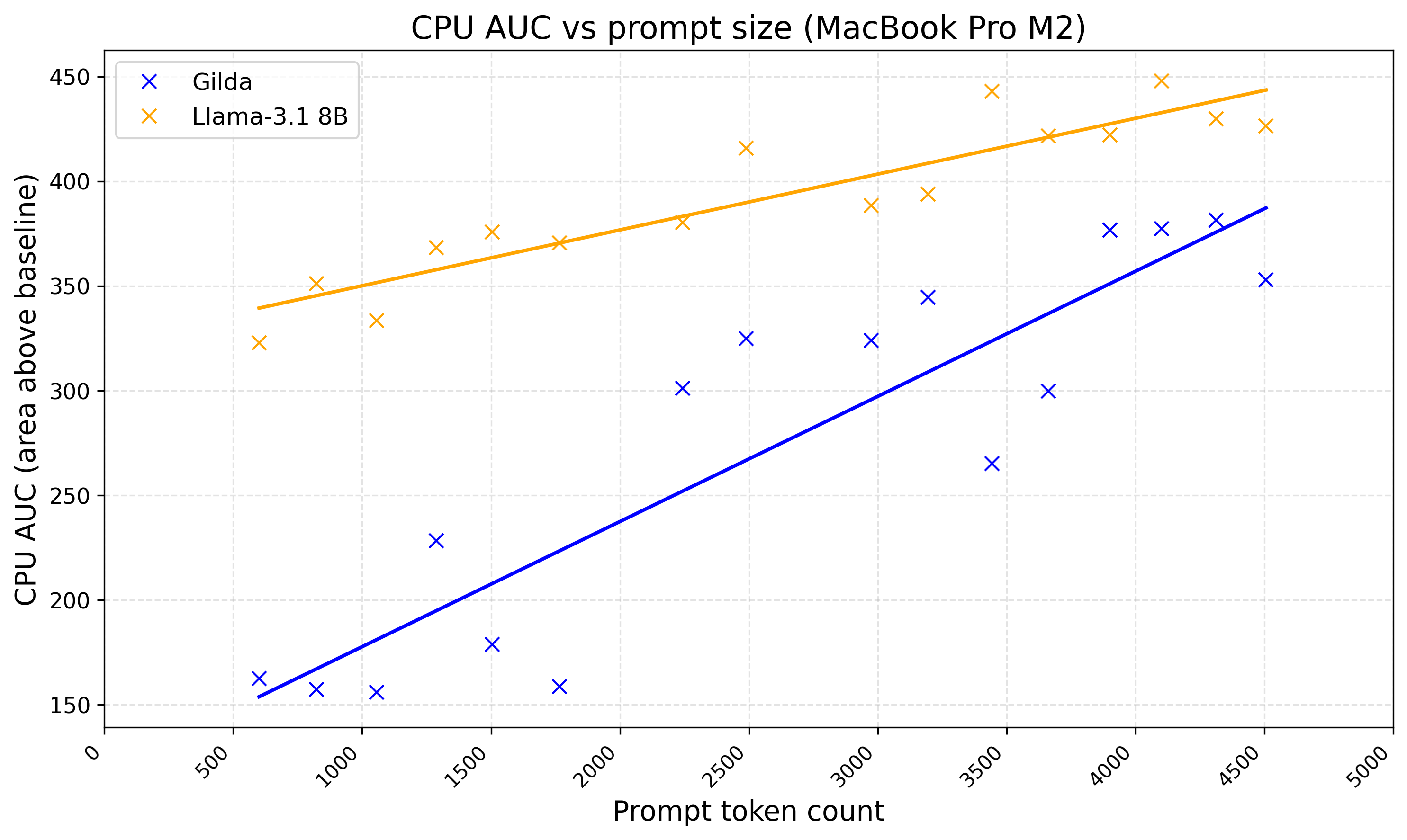}
        \caption*{(a) MacBook Pro M2}
        \label{fig:llm_scaling_m2}
    \end{subfigure}
    \hfill
    \begin{subfigure}[b]{0.98\columnwidth}
        \centering
        \includegraphics[width=\linewidth]{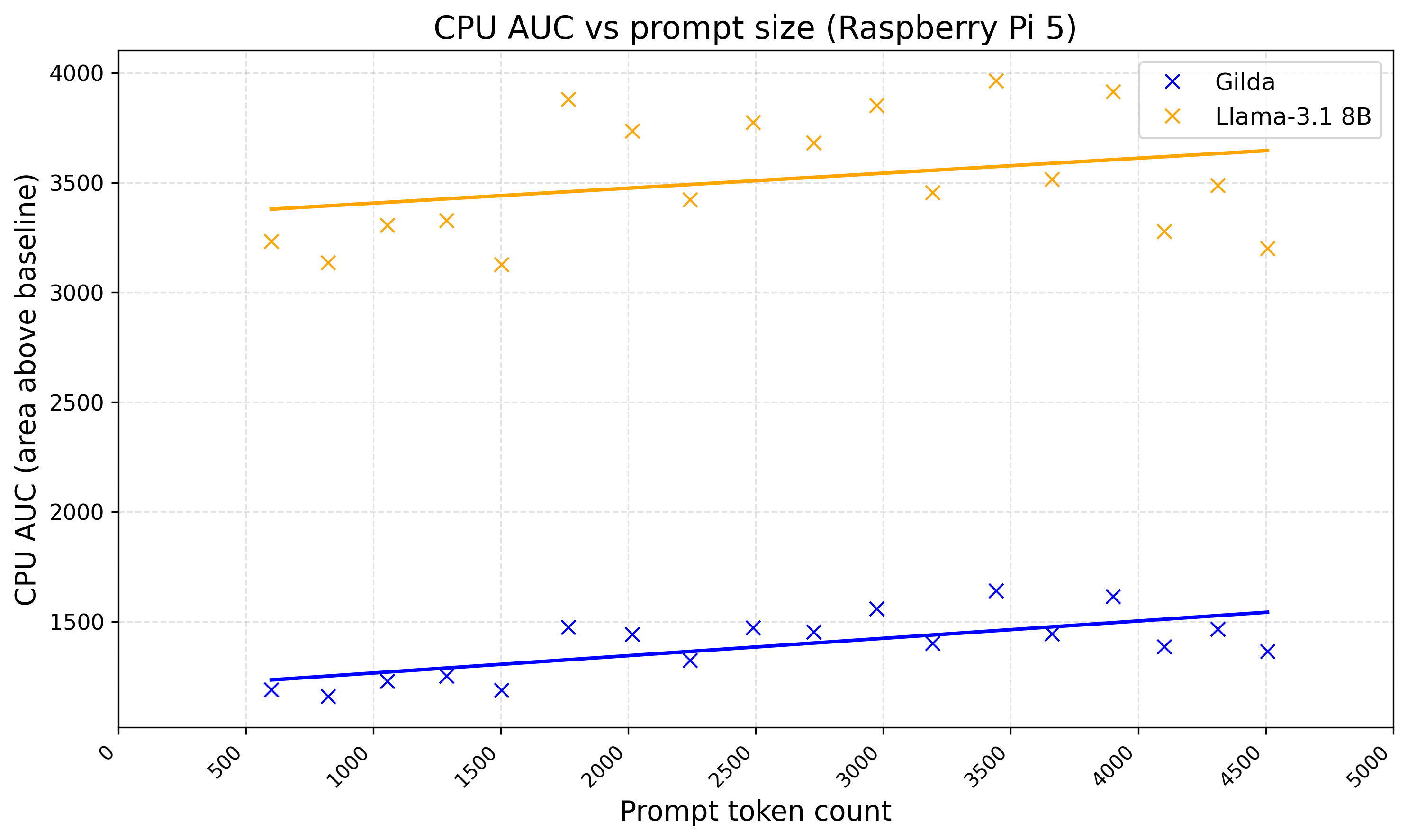}
        \caption*{(b) Raspberry Pi 5}
        \label{fig:llm_scaling_rpi}
    \end{subfigure}

    \caption{LLM CPU AUC vs.\ input token length for LLaMA-3.1 and Gilda v3.
    Panel (a) shows MacBook Pro M2 results and panel (b) shows Raspberry Pi~5
    results. Gilda reduces CPU AUC by 31.3\% and RAM AUC by 55.9\% on the
    MacBook Pro M2, and by 60.5\% (CPU) and 71.9\% (RAM) on the Raspberry
    Pi~5 at matched token lengths, while also increasing throughput (2.1$\times$
    on M2 and 2.6$\times$ on RPi5).}
    \label{fig:law1_llm_scaling}
\end{figure}

\subsection{Law 2 — VLM compute is piecewise constant with a “resolution knee”}
For VLMs, compute does not grow smoothly with image resolution. Instead, we observe a flat--then--drop pattern:
above a model-specific preprocessing clamp (1024×720 by default), CPU/RAM AUC remains constant.\\
Once the input resolution dips below this clamp, compute decreases sharply while accuracy remains unchanged.\\
This phenomenon appears in both Qwen2 and Axolotl and on both devices. It also holds after modifying the clamp:
With clamps lowered to 854×594 and 714×496, the knee shifts precisely to the new thresholds (see Fig.~\ref{fig:appendix_vlm_threshold_cpu} in Appendix~\ref{app:vlm_extended}).\\
\\

\texttt{Insight:} The VLM "knee" is not a model property---it is a preprocessing artifact. Effective pixels, not nominal pixels, determine compute. By nominal pixels we refer to the input image resolution as provided by the user (e.g., 1920×1080), whereas effective pixels denote the actual pixel budget that reaches the model after the resize-and-clamp preprocessing stage. Once the input resolution exceeds the internal clamp, all images are downsampled to the same fixed size, yielding identical effective pixels and therefore identical compute. This is empirically confirmed by the fact that lowering the clamp (1024×720 → 854×594 → 714×496) shifts the knee exactly to the new threshold in every case (see Fig.~\ref{fig:appendix_vlm_threshold_tps} and ~\ref{fig:appendix_vlm_threshold_cpu} in Appendix~\ref{app:vlm_extended}), demonstrating that compute is governed by effective pixels rather than the nominal resolution.

\begin{figure}[t]
    \centering

    \begin{subfigure}[b]{0.98\columnwidth}
        \centering
        \includegraphics[width=\linewidth]{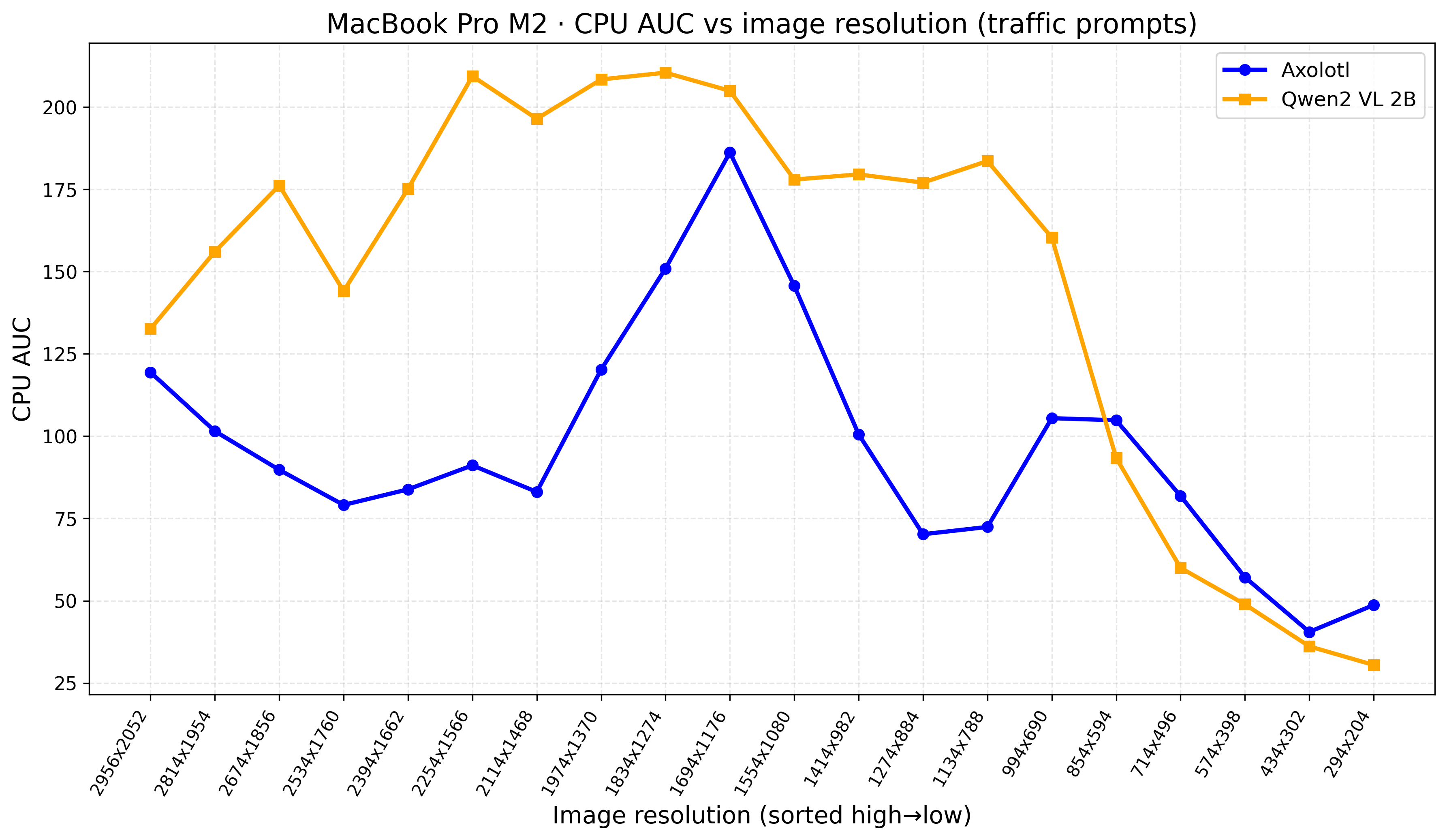}
        \caption*{(c) MacBook Pro M2}
        \label{fig:vlm_knee_m2}
    \end{subfigure}
    \hfill
    \begin{subfigure}[b]{0.98\columnwidth}
        \centering
        \includegraphics[width=\linewidth]{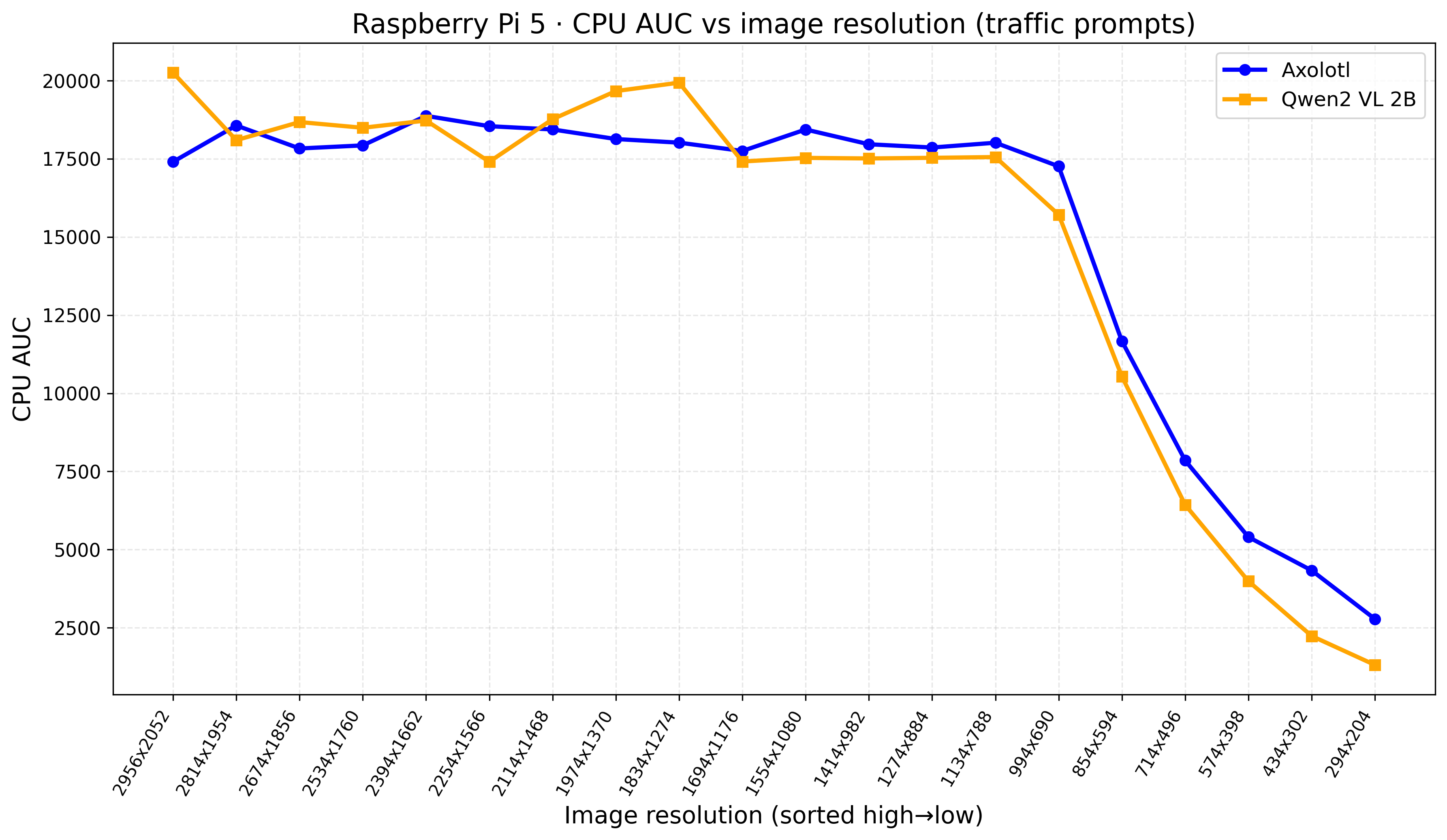}
        \caption*{(d) Raspberry Pi 5}
        \label{fig:vlm_knee_rpi}
    \end{subfigure}

    \caption{VLM CPU AUC across image resolutions for Qwen2 and Axolotl.
    Panel (c) shows MacBook Pro M2 results and panel (d) shows Raspberry Pi~5
    results. Both devices exhibit a preprocessing-induced "resolution knee":
    above the internal clamp, the two models have nearly identical AUC because
    the effective visual resolution is the same, while below the clamp compute
    drops sharply as the number of effective pixels decreases. Axolotl provides
    a modest average CPU AUC reduction (23.3\% on the MacBook Pro M2 and
    $\approx$0.2\% on the Raspberry Pi~5); the main effect illustrated here is
    the knee shape rather than the absolute gap between models.}

    \label{fig:law2_vlm_knee}
\end{figure}

\subsection*{Empirical observation}
Let $r$ be the nominal image resolution and let $\mathcal{S}_{r_{\max}}(r)$ denote the resize--and--clamp operator in the preprocessing pipeline, with width$\times$height bound $r_{\max}$. We write:
\[
u = \mathcal{S}_{r_{\max}}(r), \quad
C(r) = \tilde{C}(u), \quad
T(r) = \tilde{T}(u),
\]
where $C$ is compute (CPU/RAM AUC) and $T$ is throughput (tokens/s). Empirically, we observe that: (i) accuracy $P(r)$ remains approximately constant across all tested resolutions; (ii) $\tilde{C}$ increases with effective resolution $u$; and (iii) $\tilde{T}$ is inversely related to compute.

Under this model, if $r$ is larger than the clamp, then $u = r_{\max}$ and both $C(r)$ and $T(r)$ are flat with respect to $r$. Once $r$ falls below $r_{\max}$, the effective resolution $u$ decreases with $r$, so compute drops and throughput increases, while accuracy remains unchanged. When we lower the clamp from $1024\times 720$ to $854\times 594$ and $714\times 496$, the knee shifts to the new $r_{\max}$ in all cases (see Figs.~\ref{fig:appendix_vlm_threshold_cpu} and~\ref{fig:appendix_vlm_threshold_tps} in Appendix~\ref{app:vlm_extended}), confirming that the resolution knee is entirely induced by preprocessing rather than by the internal VLM architecture.

\subsection{Compression as an efficiency multiplier on constrained hardware}
For VLMs, the compressed Axolotl model is consistently faster:
\begin{itemize}
    \item 1.8× faster on M2
    \item 2.0× faster on RPi5
    \item Faster on 19 out of 20 (M2) and 18 out of 20 (RPi5) resolutions
\end{itemize}
CPU AUC shows moderate improvements on M2 (+23.3\% average) and near neutrality on RPi5 (0.2\% average). RAM AUC is mixed, sometimes rising when Axolotl generates longer outputs.
For VLMs, energy follows compute: Axolotl reduces per-prompt energy by 37.5\% on M2 and 5.9\% on RPi5, with similar trends over full runs. For LLMs, compression yields even stronger savings, with energy reductions exceeding 60\% on the Raspberry Pi 5 (see Tables~\ref{tab:energy_llm_all} and~\ref{tab:energy_vlm_all}).\\

\begin{table*}[t]
  \centering
  \caption{Energy consumption for LLMs across all prompts (single-run averages).}
  \begin{tabular}{llcccc}
    \toprule
    \textbf{Hardware} & \textbf{Model} & \textbf{Power (W, max)} & \textbf{Prompt Duration (s)} & \textbf{Max Wh / prompt} & \textbf{Max Wh / run} \\
    \midrule
    \multirow{2}{*}{MacBook Pro M2} 
        & LLaMA-3.1 & 43.0 & 13.0 & 0.16 & 2.95 \\
        & Gilda     & 33.6 &  8.4 & 0.08 & 1.50 \\
    \midrule
    \multirow{2}{*}{Raspberry Pi 5} 
        & LLaMA-3.1 & 14.0 & 87.8 & 0.34 & 6.50 \\
        & Gilda     & 13.6 & 34.8 & 0.13 & 2.50 \\
    \bottomrule
  \end{tabular}
  \label{tab:energy_llm_all}
\end{table*}

\begin{table*}[t]
  \centering
  \caption{Energy consumption for VLMs across all resolutions (single-run averages).}
  \begin{tabular}{llcccc}
    \toprule
    \textbf{Hardware} & \textbf{Model} & \textbf{Power (W, max)} & \textbf{Prompt Duration (s)} & \textbf{Max Wh / prompt} & \textbf{Max Wh / run} \\
    \midrule
    \multirow{2}{*}{MacBook Pro M2} 
        & Qwen2   & 40.5 & 14.0 & 0.16 & 3.20 \\
        & Axolotl & 27.1 & 13.7 & 0.10 & 2.10 \\
    \midrule
    \multirow{2}{*}{Raspberry Pi 5} 
        & Qwen2   & 15.4 & 159  & 0.68 & 13.6 \\
        & Axolotl & 14.9 & 154  & 0.64 & 12.7 \\
    \bottomrule
  \end{tabular}
  \label{tab:energy_vlm_all}
\end{table*}

\texttt{Insight:} Compression’s effect is architecture-dependent. For LLMs it reduces both compute and memory; for VLMs it primarily accelerates decoding and energy usage.

\begin{figure}[t]
    \centering

    \begin{subfigure}[b]{0.98\columnwidth}
        \centering
        \includegraphics[width=\linewidth]{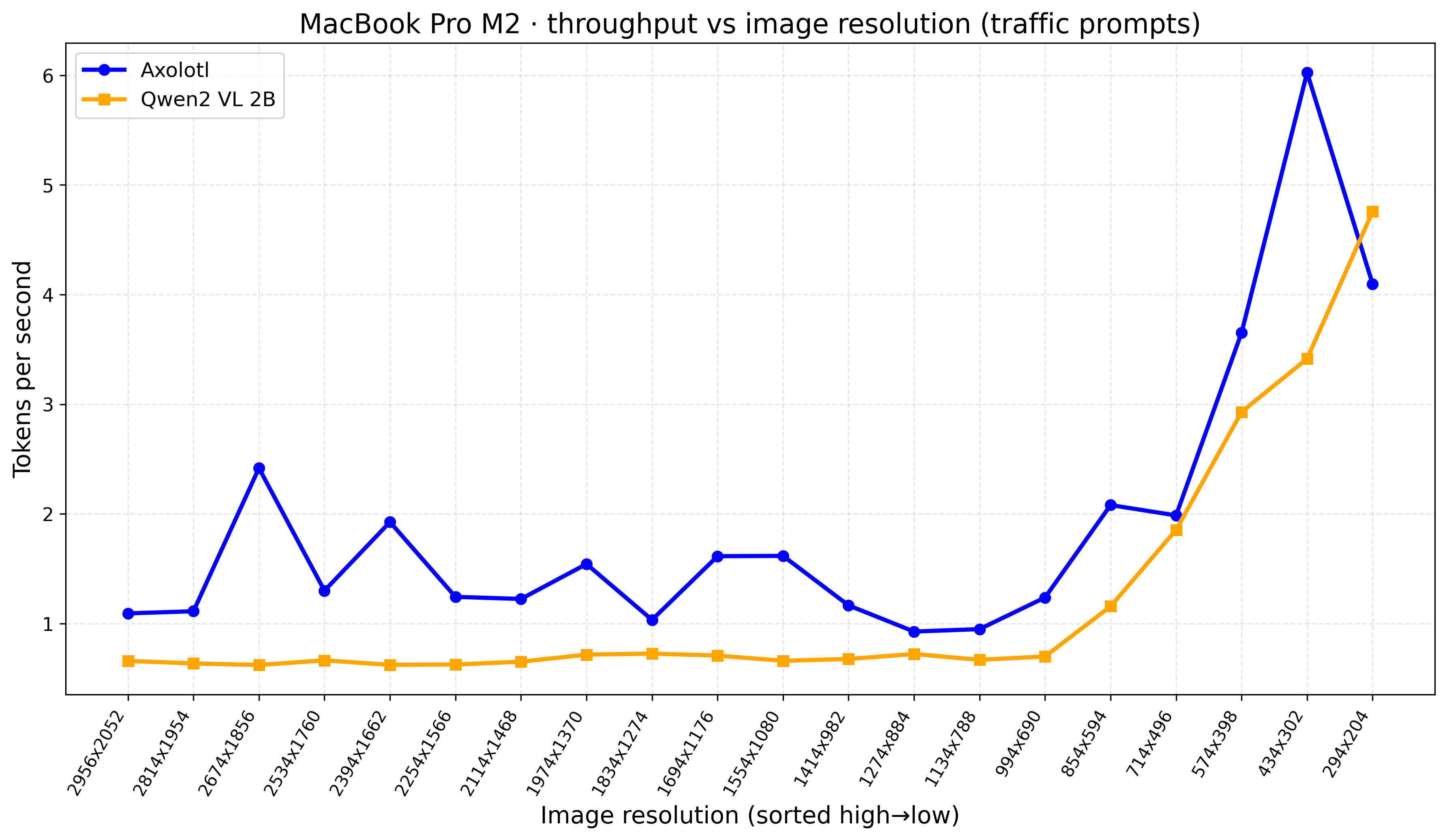}
        \caption*{(a) MacBook Pro M2}
        \label{fig:vlm_tps_m2}
    \end{subfigure}
    \hfill
    \begin{subfigure}[b]{0.98\columnwidth}
        \centering
        \includegraphics[width=\linewidth]{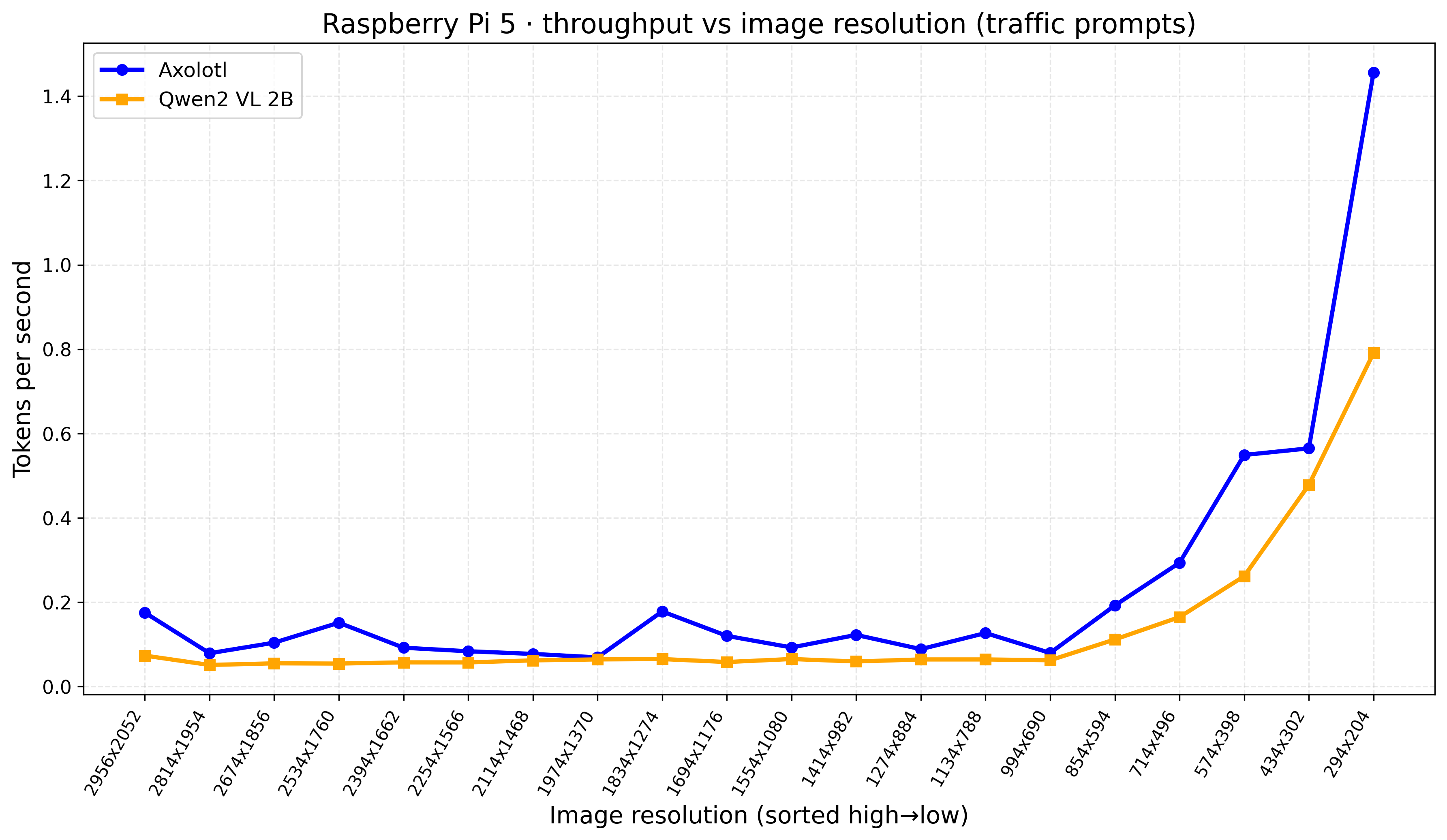}
        \caption*{(b) Raspberry Pi 5}
        \label{fig:vlm_tps_rpi}
    \end{subfigure}

    \caption{Throughput (tokens/s) for Axolotl vs.\ Qwen2 across image resolutions.
Panel (a) shows MacBook Pro M2 results and panel (b) shows Raspberry Pi~5 results.
Although the curves partially overlap at high resolutions, Axolotl achieves on
average 1.8$\times$ higher throughput on the MacBook Pro M2 and 2.0$\times$
higher throughput on the Raspberry Pi~5, and is faster on 19/20 and 18/20
resolutions, respectively.}

    \label{fig:law3_vlm_tps}
\end{figure}

\subsection{Compression preserves or improves semantic accuracy}
Across all tasks---text summarization and visual reasoning---compressed models match or exceed the accuracy of their uncompressed baselines:
\begin{itemize}
    \item Gilda: +9.1\% (M2), +13.8\% (RPi5)
    \item Axolotl: +6.9\% (M2), +5.8\% (RPi5)
\end{itemize}
This behavior aligns with reported benefits of CompactifAI compression \cite{fovet2025}: it removes redundant or noisy parameters while preserving core functionalities.\\

\texttt{Insight:} Compression functions as a form of structured regularization, improving fidelity under CPU-only constraints for local LLM and VLM workloads.

\begin{figure}[t]
    \centering

    \begin{subfigure}[b]{0.98\columnwidth}
        \centering
        \includegraphics[width=\linewidth]{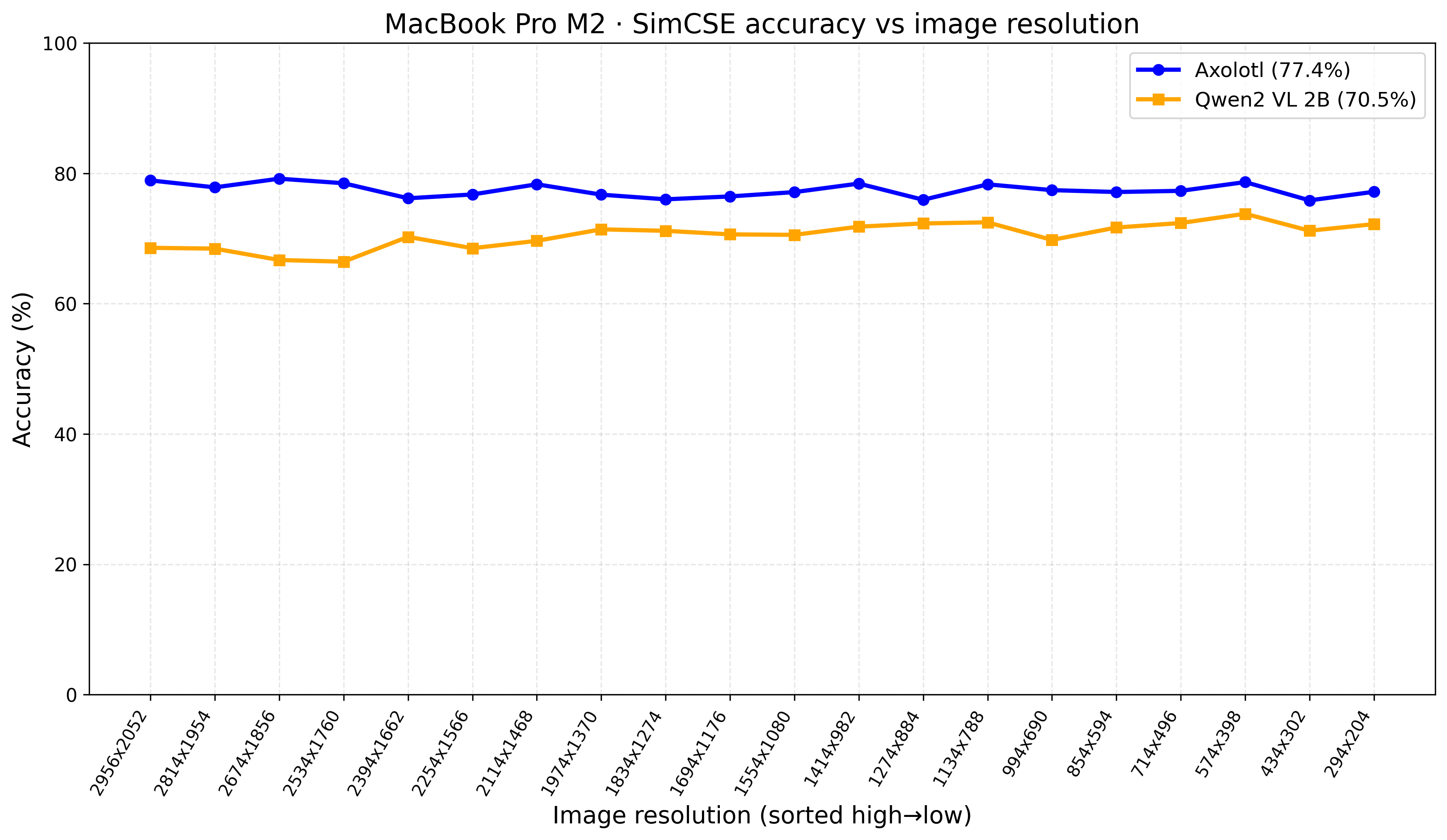}
        \caption{MacBook Pro M2}
        \label{fig:vlm_accuracy_m2}
    \end{subfigure}
    \hfill
    \begin{subfigure}[b]{0.98\columnwidth}
        \centering
        \includegraphics[width=\linewidth]{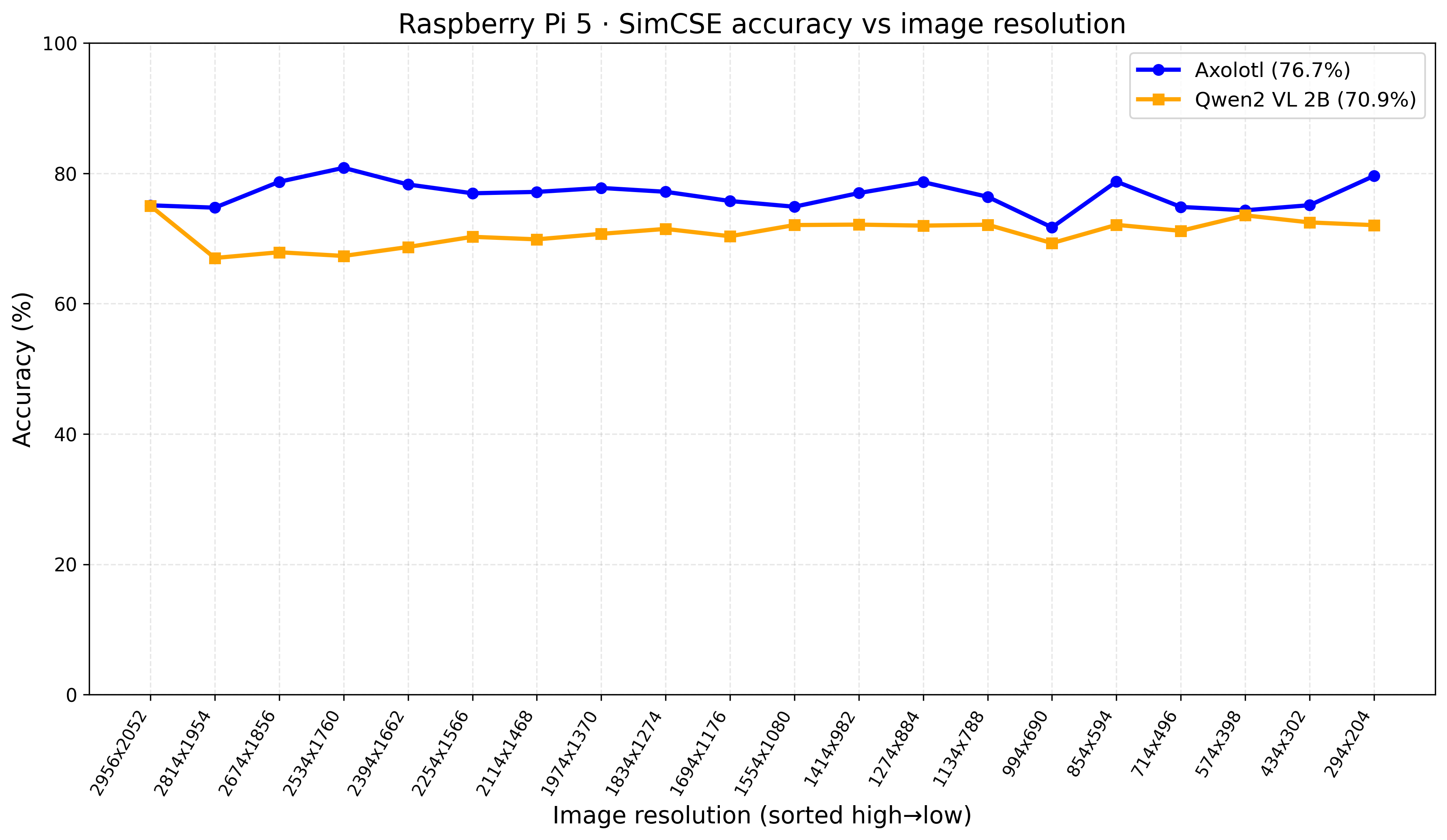}
        \caption{Raspberry Pi 5}
        \label{fig:vlm_accuracy_rpi}
    \end{subfigure}

    \caption{Semantic accuracy of Axolotl and Qwen2 across image resolutions,
    measured as SimCSE similarity to Gemini~2.5 Flash reference answers.
    Panel (a) reports MacBook Pro M2 results and panel (b) reports Raspberry
    Pi~5 results. The curves appear visually close because of the narrow y-axis
    range, but Axolotl achieves higher mean accuracy on both devices: on the
    MacBook Pro M2, average accuracy is 77.4\% for Axolotl vs.\ 70.5\% for
    Qwen2 (+6.9\%), and on the Raspberry Pi~5 it is 76.7\% vs.\ 70.9\%
    (+5.8\%).}
    \label{fig:law4_vlm_accuracy}
\end{figure}

\section{Discussion}\label{sec:cots_compatibility}

\noindent\textbf{Why LLMs scale linearly: the dominance of prefill compute.}
Decoder-only architectures process all input tokens during the prefill stage. Since output length is invariant across prompts, the dominant term in total compute is the number of input tokens processed once. This explains linear AUC scaling, monotonic throughput increases, and why compression reduces both intercept and slope. On constrained devices, reduced per-token cost yields disproportionately large efficiency gains and makes local LLM usage viable for longer contexts.

\medskip\noindent\textbf{Why VLMs exhibit a knee: preprocessing sets the effective pixel budget.}
The VLM pipeline front-loads a deterministic resize/clamp operation that produces a fixed visual embedding once resolution exceeds the clamp. Thus:
\begin{itemize}
    \item Above clamp: identical number of visual tokens $\rightarrow$ constant compute.
    \item Below clamp: fewer effective pixels $\rightarrow$ faster inference at no accuracy cost.
\end{itemize}
Changing the clamp shifts the knee, indicating that it is not an intrinsic architectural feature of the VLM. This has practical implications: controlling the preprocessing threshold is an efficiency lever that can be tuned without modifying the model, and should be treated as part of the system design for local multimodal inference.

\medskip\noindent\textbf{Why compression amplifies efficiency on edge hardware.}
Both Gilda and Axolotl are structurally reorganized to reduce parameter redundancy. This yields lower fixed overhead (important for short prompts or high-resolution images), improved cache behavior, reduced RAM pressure (especially for LLMs), and faster decoding (especially for VLMs). Because CPU-only devices lack specialized matrix-multiplication hardware, any reduction in memory bandwidth or arithmetic density produces notable savings. The effect is most pronounced on the Raspberry Pi 5, where compression turns otherwise marginal local LLM workloads into feasible ones.

\medskip\noindent\textbf{Implications for real-world local LLM deployments.}
Tokens and pixels are the fundamental units of cost. For local LLM and VLM scenarios, this suggests three design rules:
\begin{itemize}
    \item Limit prompt length and manage context as an explicit computational resource.
    \item Set image resolution below the preprocessing clamp where possible, to reduce effective visual tokens without degrading accuracy.    
    \item Use compressed models by default, especially on embedded hardware, since CompactifAI compression provides consistent reductions in CPU/RAM AUC and energy usage.
\end{itemize}
Preprocessing configuration (e.g., resolution clamps) should be documented with the same rigor as model checkpoints, as it directly shapes the system-level cost of local inference.

Energy and sustainability also matter at deployment time. Our measurements show that LLM energy drops by 50--62\% and VLM energy by 6--37\% under compression, in line with the broader push toward green AI and carbon-aware evaluation \cite{green_ai_measurement2019,ipw2025}. At scale, even modest per-call savings accumulate, especially when combined with realistic electricity prices such as those reported by Eurostat for European households \cite{eurostat_electricity_prices_2024}. Taken together, the results on MacBook Pro M2 and Raspberry Pi 5 indicate that, with appropriate compression and preprocessing, local LLMs can operate efficiently across a wide range of CPU-only devices, from high-end laptops to low-cost edge boards.

\medskip\noindent\textbf{Limitations and future work.}
The present study has several limitations. Power measurement is based on coarse external meters rather than board-level instruments. Our accuracy metric relies on semantic similarity rather than human-validated benchmarks, and the visual evaluation uses a single image. Future work includes evaluating multi-image and multi-task VLMs, assessing concurrency and batching behavior under multi-user loads, exploring carbon-aware scheduling for local LLM services, and generalizing threshold-based optimization to other preprocessors and runtimes.

\section{Conclusions} \label{sec:conclusions}

This work presented a cross-device, cross-modality study of CPU-only inference for large language models and vision--language models in the context of local deployment. By systematically varying input complexity and applying a unified AUC-based profiling methodology, we uncovered two empirical scaling laws: (i) LLM compute scales approximately linearly with token length, and (ii) VLM compute is piecewise constant with a preprocessing-induced "resolution knee", remaining flat above an internal clamp and decreasing sharply below it. These behaviors were consistent across both a MacBook Pro M2, representative of laptop-class local LLM usage, and a Raspberry~Pi~5, representative of constrained embedded deployments.

On top of these scaling laws, we showed that CompactifAI compression is not merely a memory-saving technique but an effective efficiency multiplier for CPU-only inference. The compressed LLM (Gilda) reduced CPU and RAM AUC by up to 60.5\% and 71.9\% on the Raspberry~Pi~5, while simultaneously delivering 2.6$\times$ higher throughput and up to 17.9\% higher semantic accuracy. The compressed VLM (Axolotl) roughly doubled throughput on both devices, reduced CPU AUC on the M2, and preserved or improved semantic accuracy across all tested resolutions. Energy consumption followed these trends, with up to 62\% reductions for LLMs and substantial savings for VLMs, indicating that compression and preprocessing are practical levers for sustainable local inference on CPUs.

The observation that the VLM resolution knee is entirely governed by the preprocessing clamp---and can be shifted without accuracy loss---has direct implications for model deployment. It implies that image resolution should be treated as a first-class resource knob, alongside token budgets, and that preprocessing configuration must be documented as part of the model specification for local LLM and VLM systems. More broadly, our results support three actionable principles for practitioners: (i) manage tokens and pixels explicitly as cost drivers; (ii) deploy compressed models by default, especially on resource-constrained hardware; and (iii) monitor energy (Wh per prompt or per run) as a core performance metric.

Looking forward, it will be important to validate these findings on broader task suites and datasets, to include human-verified accuracy benchmarks, and to explore concurrency and batching effects under realistic multi-user workloads. Extending the analysis to additional runtimes and hardware, and integrating carbon-aware scheduling into edge deployment pipelines, are promising directions. Nonetheless, the present study already indicates that compression techniques such as CompactifAI, combined with careful system-level measurement, can significantly expand the feasibility of local LLM and VLM deployments on CPU-only devices.

\section{Acknowledgement}
We also thank \textbf{Laura Cabello} for her guidance and expert advice, and \textbf{Marlon Cajamarca Vega} for his valuable technical input and support during the experiments.

\appendix

\section{LLM Extended Results}
\label{app:llm_extended}

\subsection{LLM Accuracy Analysis}
\begin{figure}[hbtp]
    \centering
    \begin{subfigure}[b]{0.98\columnwidth}
        \centering
        \includegraphics[width=\linewidth]{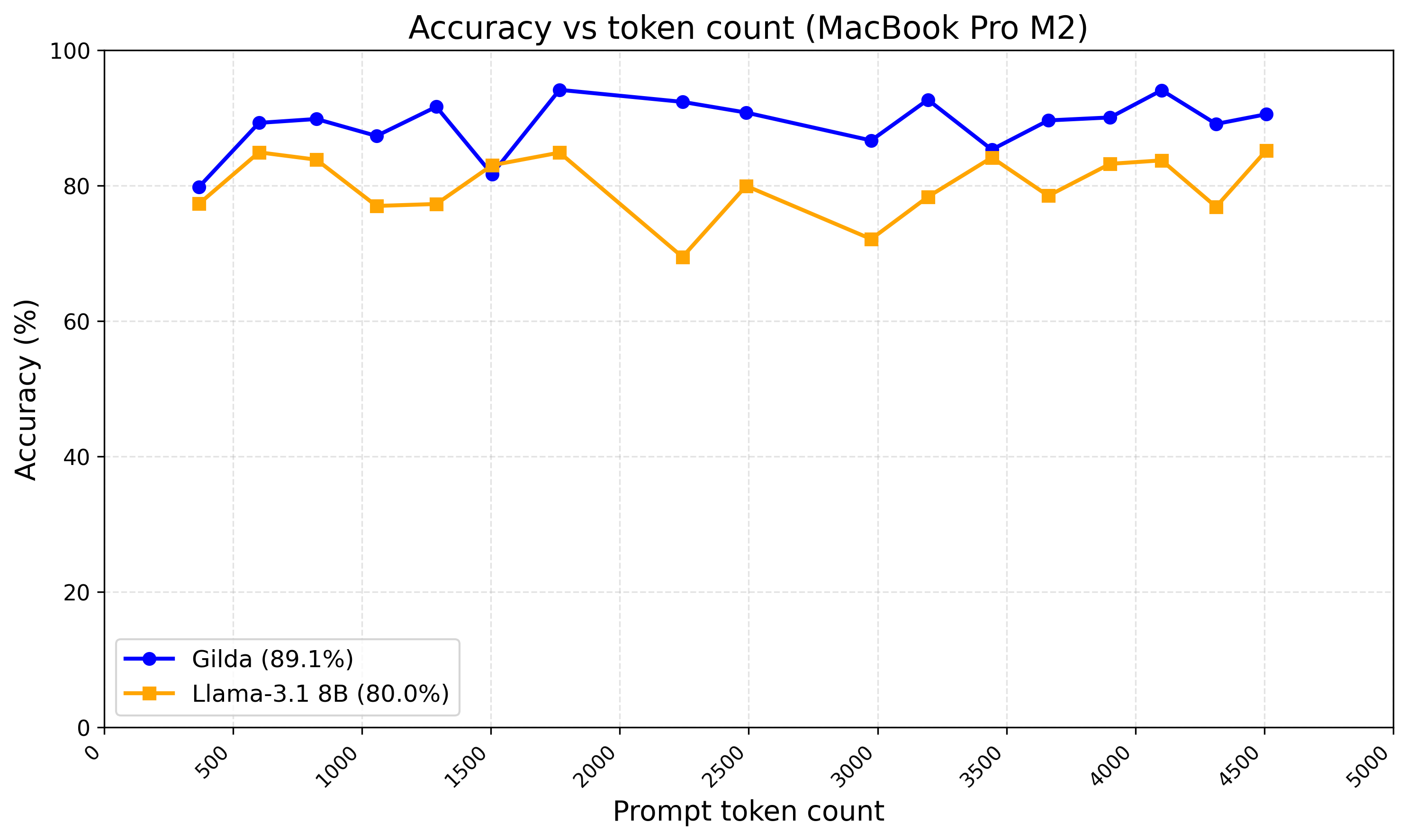}
        \caption*{(a) MacBook Pro M2}
    \end{subfigure}
    \hfill
    \begin{subfigure}[b]{0.98\columnwidth}
        \centering
        \includegraphics[width=\linewidth]{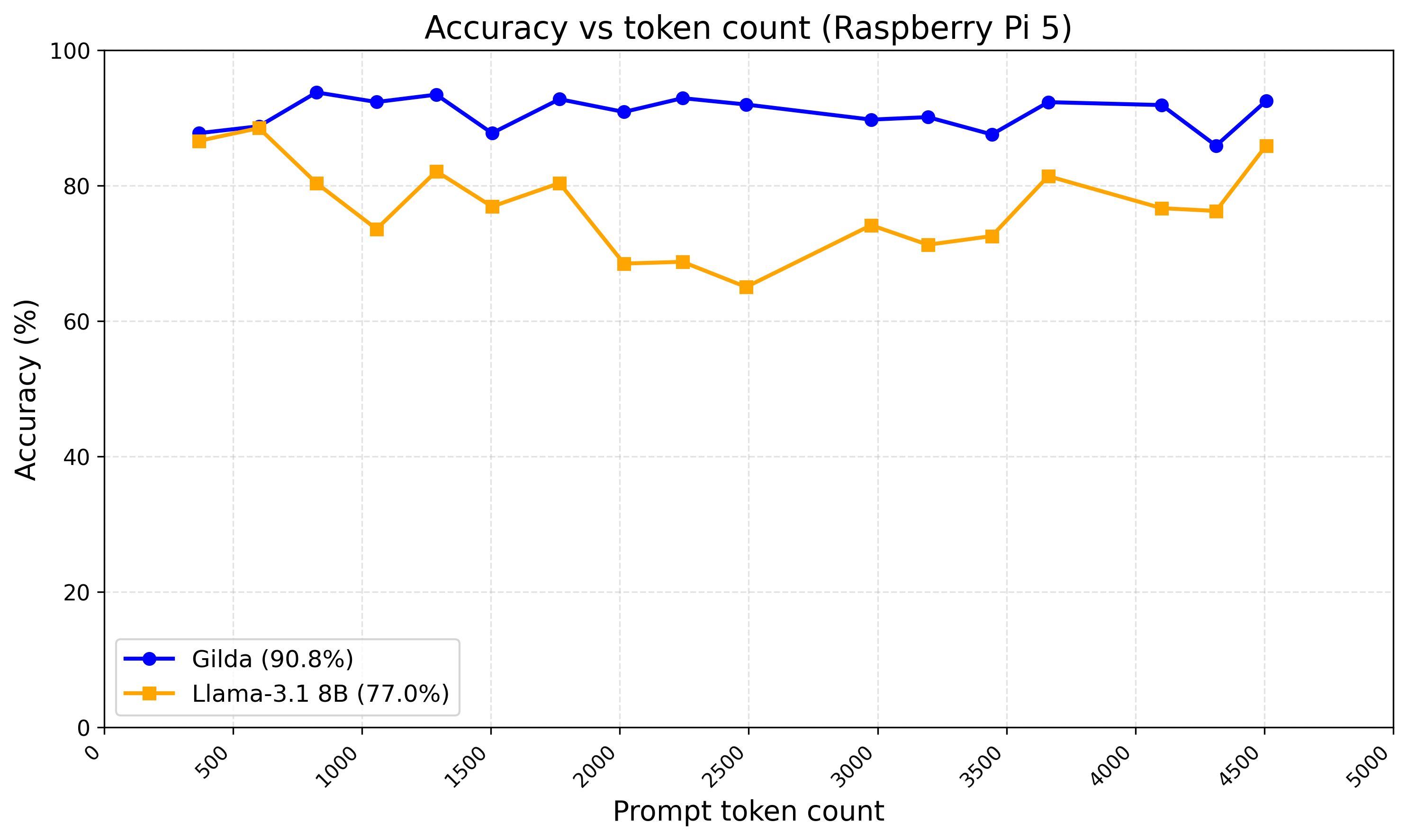}
        \caption*{(b) Raspberry Pi 5}
    \end{subfigure}
    \caption{Semantic accuracy of Gilda v3 and LLaMA-3.1 across the 19 LLM
    prompts, measured as SimCSE similarity to Gemini~2.5 Flash references.
    Panel (a) shows MacBook Pro M2 results and panel (b) shows Raspberry
    Pi~5 results. On the MacBook Pro M2, Gilda attains an average accuracy
    of 89.1\% compared to 80.0\% for LLaMA-3.1 (+9.1\%). On the Raspberry
    Pi~5, the averages are 90.8\% vs.\ 77.0\% (+13.8\%).}
    \label{fig:appendix_llm_accuracy}
\end{figure}

Figure~\ref{fig:appendix_llm_accuracy} shows the SimCSE similarity
scores against the Gemini~2.5 Flash references for all 19 prompts.
As described in Section~\ref{sec:results}, we apply symmetric outlier 
removal whenever one of the two models produces degenerate or trivial outputs.

\FloatBarrier

\section{VLM Extended Results}
\label{app:vlm_extended}

\newpage
\subsection{Resolution-threshold curves (854×594, 714×496)}
\begin{figure}[t]
    \centering
    \begin{subfigure}[b]{0.98\columnwidth}
        \centering
        \includegraphics[width=\linewidth]{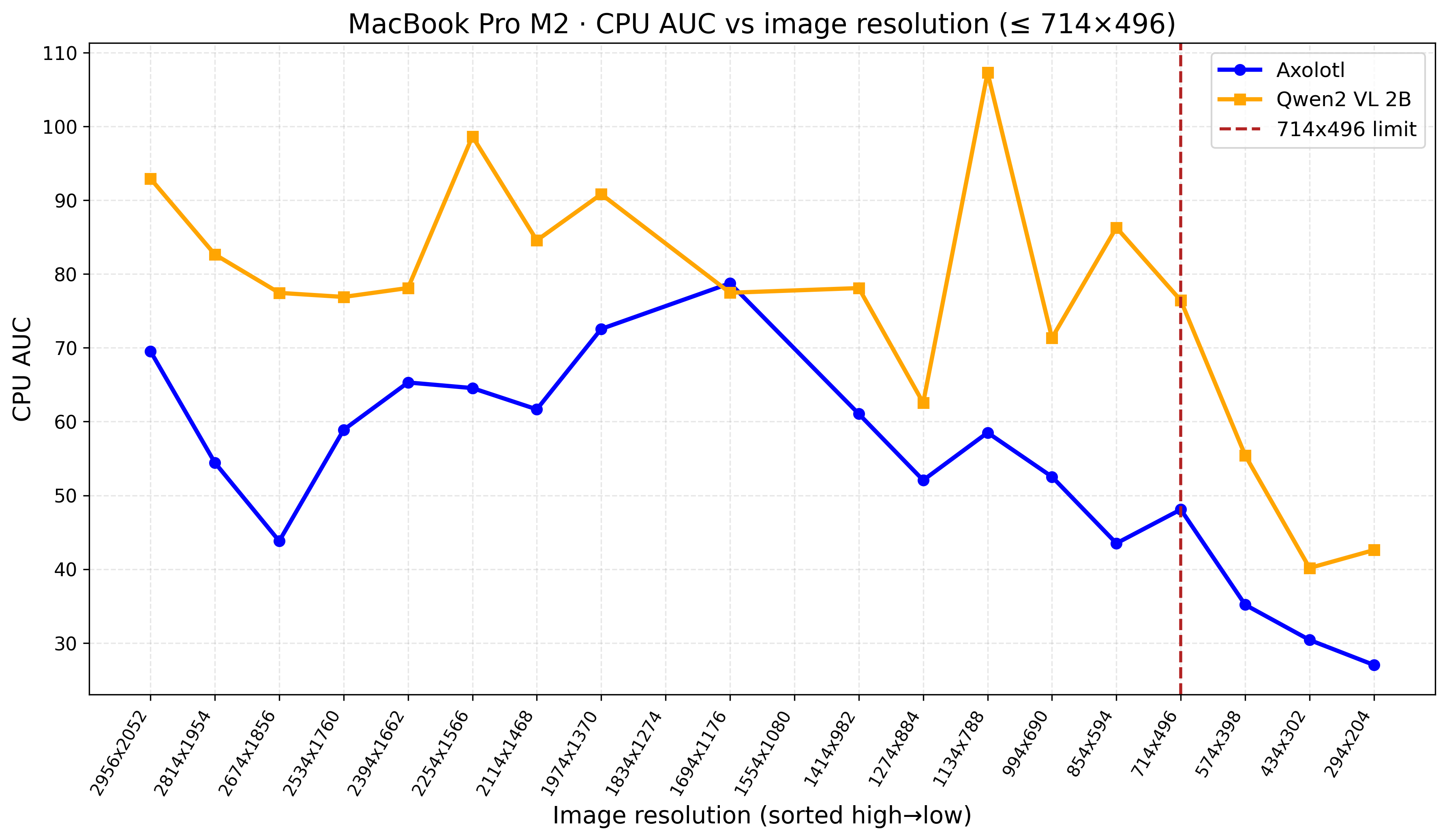}
        \caption*{(a) 714×496 threshold}
    \end{subfigure}
    \hfill
    \begin{subfigure}[b]{0.98\columnwidth}
        \centering
        \includegraphics[width=\linewidth]{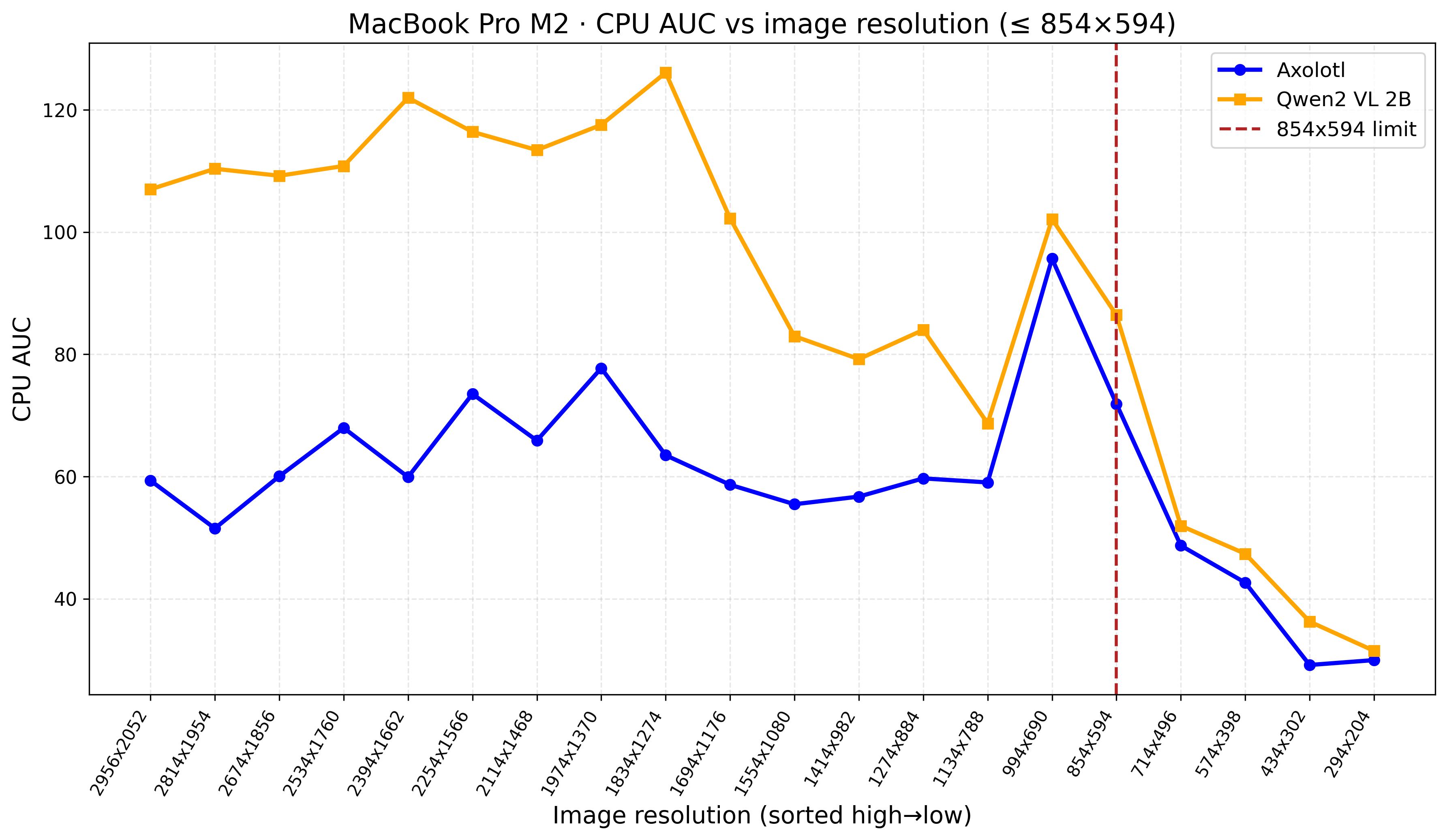}
        \caption*{(b) 854×594 threshold}
    \end{subfigure}
    \caption{CPU AUC comparison across the two new preprocessing thresholds
for Qwen2-VL-2B and Axolotl on Macbook Pro M2. Panel (a) shows results for the 714$\times$496
threshold and panel (b) for the 854$\times$594 threshold. In both cases the
"resolution knee" shifts to the new clamp value, confirming that the knee
is induced by the resize--and--clamp preprocessing rather than by the
internal VLM architecture.}

    \label{fig:appendix_vlm_threshold_cpu}
\end{figure}

Figure~\ref{fig:appendix_vlm_threshold_cpu} shows full CPU AUC curves
for the additional preprocessing clamps used in the theoretical validation
of the resolution knee.

\begin{figure}[hbtp]
    \centering
    \begin{subfigure}[b]{0.7\columnwidth}
        \centering
        \includegraphics[width=\linewidth]{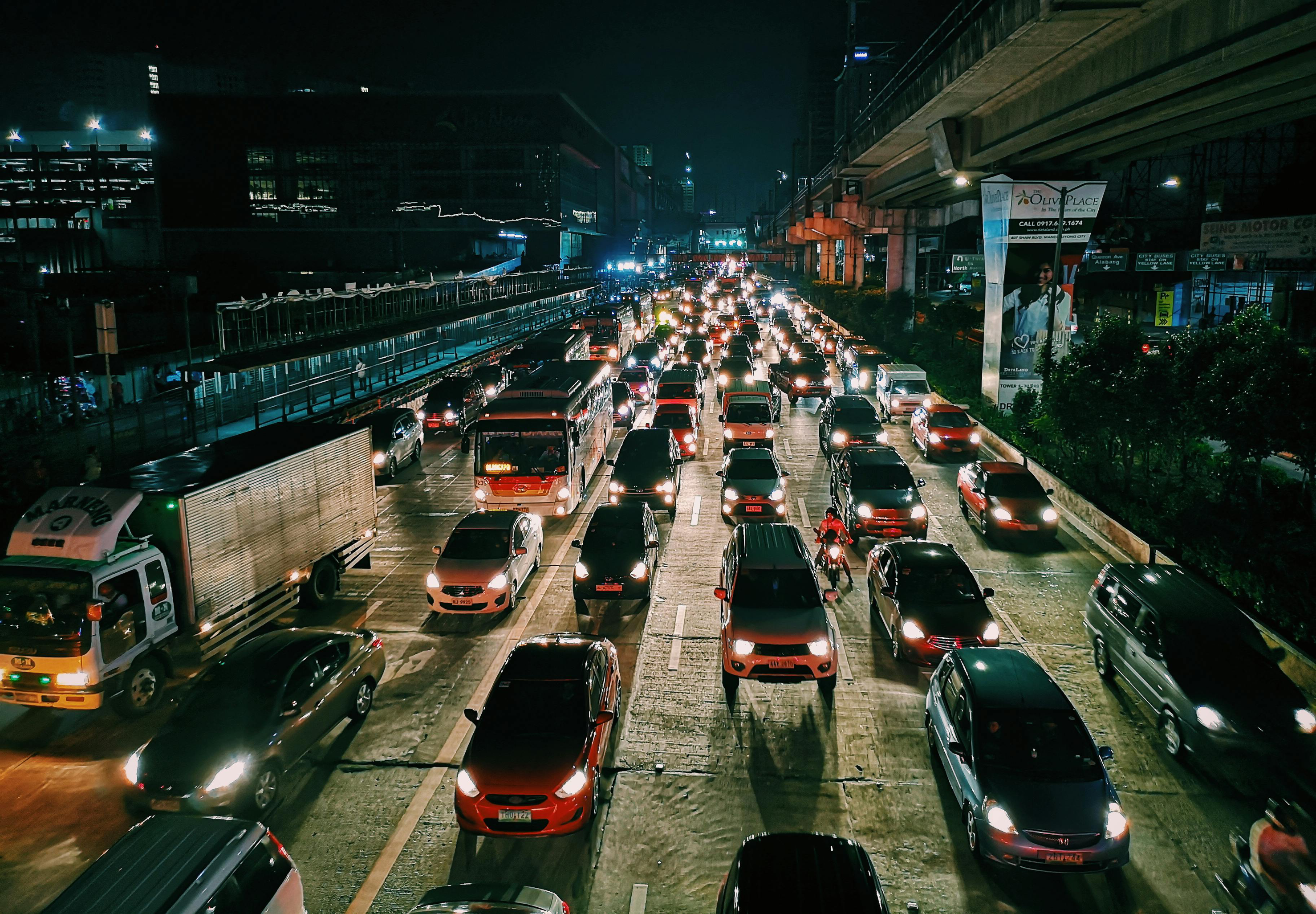}
    \end{subfigure}
    \hfill
    \caption{Multi-entity traffic image used for the VLM experiments.}
    \label{fig:traffic}
\end{figure}

Figure~\ref{fig:traffic} is the multi-entity traffic image used in our experiments.

\subsection{Throughput curves for all clamps}
\begin{figure}[t]
    \centering
    \begin{subfigure}[b]{0.98\columnwidth}
        \centering
        \includegraphics[width=\linewidth]{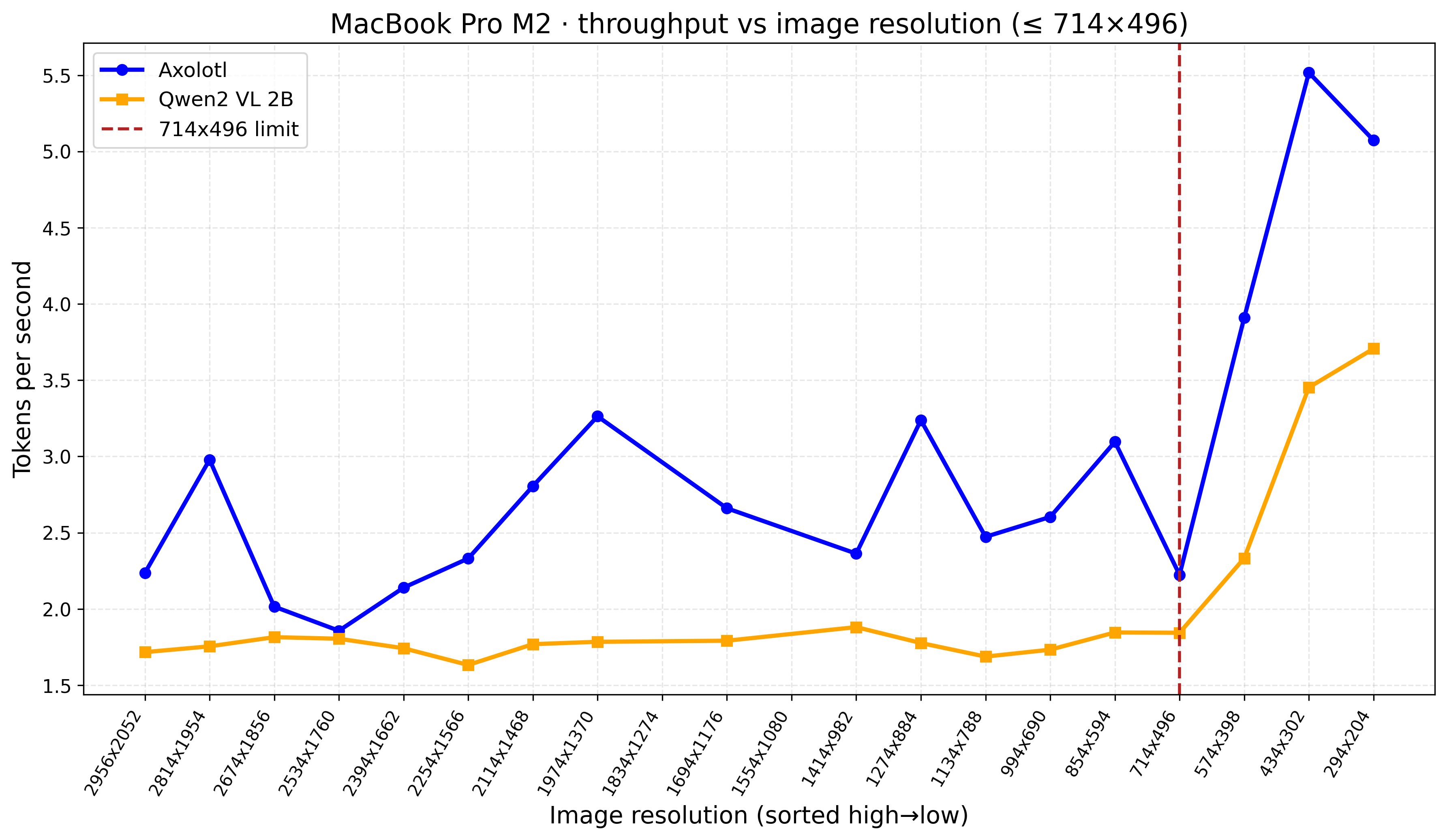}
        \caption*{(c) 714×496 threshold}
    \end{subfigure}
    \hfill
    \begin{subfigure}[b]{0.98\columnwidth}
        \centering
        \includegraphics[width=\linewidth]{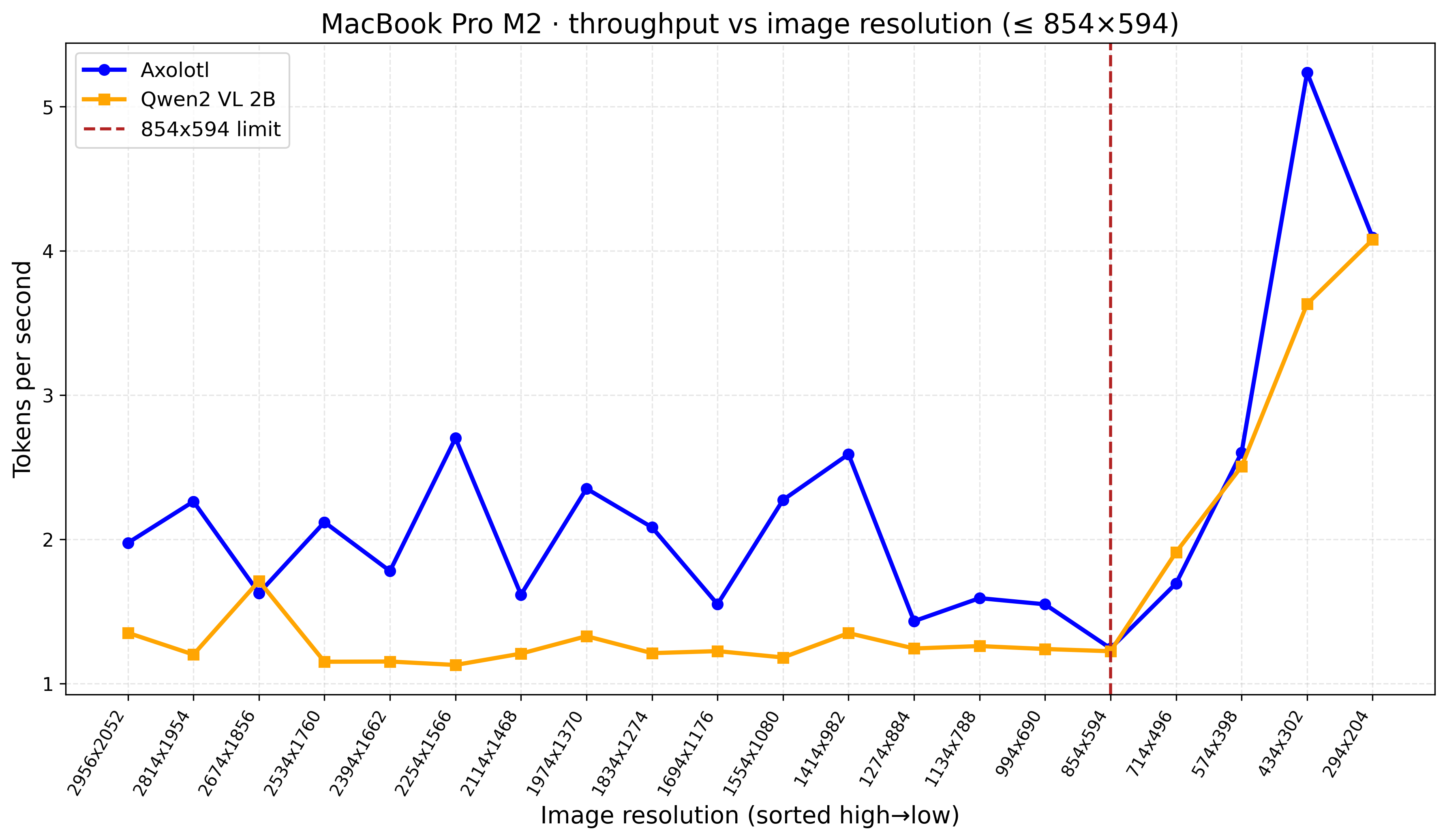}
        \caption*{(d) 854×594 threshold}
    \end{subfigure}
    \caption{Throughput comparison across the two new preprocessing thresholds
for Qwen2-VL-2B and Axolotl on Macbook Pro M2. Panel (c) shows the 714$\times$496 clamp and
panel (d) the 854$\times$594 clamp. Lower clamps translate into higher
throughput once the nominal resolution falls below the clamp, while
preserving the qualitative advantage of the compressed model.}

    \label{fig:appendix_vlm_threshold_tps}
\end{figure}

Figure~\ref{fig:appendix_vlm_threshold_tps} reports the corresponding throughput
(tokens/s) curves for the same preprocessing thresholds, confirming that
lower clamps translate into higher throughput without accuracy degradation.

\FloatBarrier
\printbibliography

\end{document}